\documentclass[letterpaper, 10 pt, conference]{ieeeconf}

\IEEEoverridecommandlockouts
\usepackage[utf8]{inputenc}
\usepackage{graphicx}
\usepackage{amsmath, amssymb}

\let\labelindent\relax

\usepackage{enumitem}
\usepackage{siunitx}
\usepackage{booktabs}
\usepackage{placeins} 
\usepackage{cite}
\usepackage[hidelinks]{hyperref}

\setlist{itemsep=4pt, topsep=4pt, leftmargin=*, labelsep=4pt}

\title{\LARGE \bf
Closing the Loop in Teleoperation: Episode-Level Data Quality Assessment and Feedback for High-Quality Demonstration Collection
}

\author{%
  Gokul Narayanan\textsuperscript{*}, Yash Shahapurkar\textsuperscript{*}, Melih Erdogan, Brian Zhu, Eugen Solowjow%
}

\begin{document}
\maketitle

\begingroup
\renewcommand\thefootnote{}%

\footnotetext{\footnotesize
\thanks{$^{*}$ These authors contributed equally to this work.}

\thanks\texttt{\{gokul.sathya\_narayanan, yash.shahapurkar, melih.erdogan, brian.zhu, eugen.solowjow\}@siemens.com} are with Siemens Corporation.\\
}%
\addtocounter{footnote}{-1}%
\endgroup

\thispagestyle{empty}
\pagestyle{empty}

\begin{abstract}
%
Industrial automation is at a pivotal moment, as Physical AI is driving a transition from rigid, hand-engineered automation systems toward more flexible and adaptive systems.
This shift has created a growing demand for large-scale, real-world robot demonstration data, making teleoperation an increasingly important mechanism for data collection.
However, high-quality teleoperated demonstrations remain difficult to obtain in practice, as novice operators often produce episodes that are task-successful but suboptimal for downstream use due to inefficient motion, repeated corrections, or operation near robot joint limits. 
We present a Data Quality Assessment and Feedback (DQAF) framework that closes the loop in teleoperation by providing immediate post-episode feedback grounded in semantic task progress and robot telemetry.

The framework extracts quality relevant signals such as sub-task progress, motion smoothness, stalls, kinematic limits and converts them into structured quality assessments and actionable natural-language feedback. Unlike binary success/failure feedback, the proposed system explains why an episode is suboptimal and highlights specific behaviors to correct in the next trial. 
We evaluate the framework through a diagnostic validation study and a pilot user study.
In the validation study, the system is compared with a human reviewer during dataset curation, producing rejection reasons and actionable feedback for improvement.
In the pilot study with three novice operators across two manipulation tasks, the operator who received the system’s immediate, automated post-episode feedback improved faster than those who did not, producing higher-quality demonstrations sooner.
\end{abstract}

\section{INTRODUCTION}
Vision-Language-Action (VLA)  models are emerging as a key enabler of Physical AI, offering a path toward more flexible and adaptive automation on the factory floor.
Recent progress~\cite{openvla2024,Groot2025,Pi02025,Pi052025} in  this field has increased the demand for large, diverse, high-quality robot manipulation datasets.
Among data collection strategies, teleoperation remains a common and effective approach because state-of-the-art methods often learn best from demonstrations collected directly on the physical system, where the fine-grained structure of real-world interaction, embodiment, and sensing is preserved. 
In practice, however, the usefulness of teleoperated data depends not only on task completion, but also on execution quality. Demonstrations that are jerky, inconsistent, or close to joint limits can degrade downstream policy learning~\cite{sakr2024}, even when they are nominally successful.

This challenge is especially pronounced for novice teleoperators. While they are typically given instructions or example demonstrations that convey what a good trajectory should look like, consistently reproducing such behavior during live teleoperation remains difficult. Operators must simultaneously cope with high-dimensional control, imperfect viewpoints, reduced depth cues, and the cognitive burden of coordinating many degrees of freedom.
Consequently, even when the desired behavior is understood conceptually, inexperience and cognitive overload~\cite{Li2025} can prevent novices from producing reliable, high-quality demonstrations. These challenges are further amplified by the lack of explicit, post episode feedback on demonstration quality. In many cases, feedback is available only at the level of task success or failure, with little guidance about demonstration quality.
As a result, an operator may complete a task successfully yet still produce data that is suboptimal for policy training and is only identified later during data curation.
This creates a costly gap between data collection and validation, where poor-quality episodes are discovered only after significant operator time has been spent.

In this work, we treat teleoperation as a closed-loop data-collection process rather than a passive recording stage. 
To this end, we introduce a lightweight Data Quality Assessment and Feedback (DQAF) framework that evaluates each teleoperated episode immediately after completion and provides actionable natural-language feedback to the operator. 
This design is motivated in part by prior findings that novice users often hold inaccurate assumptions about what demonstrations are most useful, and that interactive guidance can significantly improve teaching efficiency as well as robot learning and generalization performance~\cite{sakr2025}. 

Our framework leverages semantic task progress and robot telemetry grounded in task context and expert references to assess episode quality and deliver targeted post-episode guidance to the operator beyond binary success/failure.
In doing so, it aims to improve demonstration quality at the source, reduce the effort required to train novice teleoperators, and lessen downstream data-curation burden. As teleoperation becomes increasingly important for scaling data collection for modern robot foundation models, tools of this kind can help broaden participation in data collection while increasing the usefulness of the resulting demonstration data.

We evaluate the approach on teleoperating a Unitree G1 humanoid robot using a Meta Quest-based hand-tracking interface. The evaluation includes diagnostic validation against expert review and a pilot study with novice teleoperators. Together, these experiments provide initial evidence that the framework can identify suboptimal demonstrations and may support faster operator improvement during data collection.

The main contributions of this work are as follows:
\begin{itemize}
    \item A \emph{multimodal} framework for  assessment of  demonstration data quality from semantic task progress, telemetry, task context, and expert reference demonstrations.
    \item An AI-assisted system that translates assessment results into \emph{natural language feedback} for the teleoperator, grounded on the assesment.
    \item A two-stage evaluation with diagnostic validation and a pilot user study, providing preliminary evidence of both framework reliability and the utility of immediate post-episode feedback.
\end{itemize}
\section{RELATED WORK}
\paragraph{Teleoperator performance, training, and interface support}
Teleoperation performance depends strongly on operator expertise, perceptual cues, and interface design~\cite{aoki2026}. Compared to experts, novice operators typically move more slowly, commit more errors, and produce less stable demonstrations~\cite{tugal2025}. 
These difficulties are often amplified by limited depth perception, camera-view misalignment, and reduced situational awareness, leading to corrective oscillations, misgrasps, and stalled execution~\cite{sakr2018}.
Related findings in remote space robotics likewise show that delayed or incomplete visual feedback degrades manipulation performance under constrained viewpoints~\cite{kazanzides2021}. Prior work has therefore explored improved teleoperation interfaces~\cite{akgun2012,fang 2025} including actuated camera systems and immersive viewpoints, to provide better depth cues, reduce cognitive load, and improve operator performance~\cite{actuatedneck2024}.
These works primarily focus were on improving teleoperation interfaces to obtain higher-quality demonstration data from operators.

\paragraph{Demonstration quality and dataset curation}
As robot-learning datasets scale, teleoperation enables diverse real-robot interaction but also introduces substantial cross-operator variability. 
RoboTurk demonstrated large-scale crowd-sourced teleoperation while highlighting the inconsistency of novice demonstrations~\cite{roboturk2018}, and BridgeData V2 aggregates over 60{,}000 trajectories while still containing suboptimal trajectories, underscoring the difficulty of maintaining consistent quality at scale~\cite{bridgedata2023}. 
To reduce the burden of manual review, recent work has focused on automated post-hoc curation. SCIZOR prunes low-value transitions using self-supervised progress prediction and redundancy filtering~\cite{scizor2026}, CUPID ranks demonstrations according to their causal effect on downstream closed-loop performance~\cite{cupid2025}, and DataMIL estimates trajectory influence to prioritize data that most improves task learning~\cite{datamil2026}.
Practical tools such as HuggingFace's \texttt{score\_lerobot\_episodes} similarly operationalize episodic checks for visual clarity, motion artifacts, robot health, and idle time~\cite{huggingfacescore2025}. 
Collectively, this line of work shows the need for scalable quality assessment, but primarily addresses data filtering after collection rather than helping operators improve while data are being gathered.

\paragraph{Automated feedback during demonstration collection}
Existing approaches on this topic largely focus on pre-task instruction, interface-level support, or after-trial scoring and replay.
For example,~\cite{Phaijit2023} investigate user-interface interventions for improving learning from demonstration, and~\cite{Jiahao} show that mixed-reality visualizations can provide useful feedback during teleoperation and kinesthetic teaching. 
This work~\cite{Dall2025} further notes that feedback after each trial is helpful for helping users assess demonstration quality, although their feedback is primarily score and replay-based rather than diagnostic.
In contrast to prior approaches that focus on operator training, interface design, or post-hoc dataset filtering, our work targets immediate post episode natural language feedback to the teleoperator.


\section{Problem Formulation and Requirements}
\label{sec:problem_formulation}

We model a teleoperation episode as a synchronized multimodal trajectory
\begin{equation}
\tau=\{(I_t,s_t,a_t)\}_{t=1}^{T}
\end{equation}
where \(I_t\) is the camera observation, \(s_t\) is the robot proprioceptive state, and \(a_t\) is the operator action (teleoperation command) at time \(t\).

Given the multimodal trajectory, our objective is to produce three coupled outputs:
\begin{equation}
\mathcal{O}(\tau)=\bigl(q(\tau),\,\mathcal{E}(\tau),\,\mathcal{F}(\tau)\bigr)
\end{equation}
where \(q(\tau)\in[0,10]\) is episode quality, \(\mathcal{E}(\tau)\) is a time-aligned semantic-telemetry state trace, and \(\mathcal{F}(\tau)\) is actionable natural-language operator feedback.

The DQAF system $\mathcal{O}(\tau)$ must jointly capture the following to provide natural language feedback to the teleoperator:
\begin{itemize}
\item \textbf{Task Completion}:  successful completion of the given task
\item \textbf{Task Progression}: subtask completion and progress
\item \textbf{Execution quality}: smooth, stable, and optimal control

\end{itemize}

\subsection{System Requirements}
\textbf{R1 (Actionable feedback):} The system must provide immediate post-episode feedback that states what failed, when it failed (time/subtask), and how to correct behavior.

\textbf{R2 (Semantic control coupling):} Quality evaluation must jointly use high-level semantic progression and low-level telemetry; semantically successful but suboptimally executed episodes (and vice versa) must be downgraded.

\textbf{R3 (Robustness to transients):} Isolated short deviations should not dominate episode judgment; persistent/repeated violations should be penalized.

\textbf{R4 (Curation compatibility):} Outputs must support reproducible filtering, ranking, and downstream data curation.

\textbf{R5 (Auditability):} Each quality decision must be traceable to metric values, thresholds, and time-localized semantic evidence.

\subsection{Assumptions}
\textbf{A1 (Reference demonstration availability).} Each task requires a set of captioned images $x^{\mathrm{ref}}=\{(I_t^\star,c_t^\star)\}_{t=1}^R$ sampled from a high-quality expert trajectory to anchor expected execution quality.


\begin{figure*}[t]
    \centering
    \includegraphics[width=\linewidth]{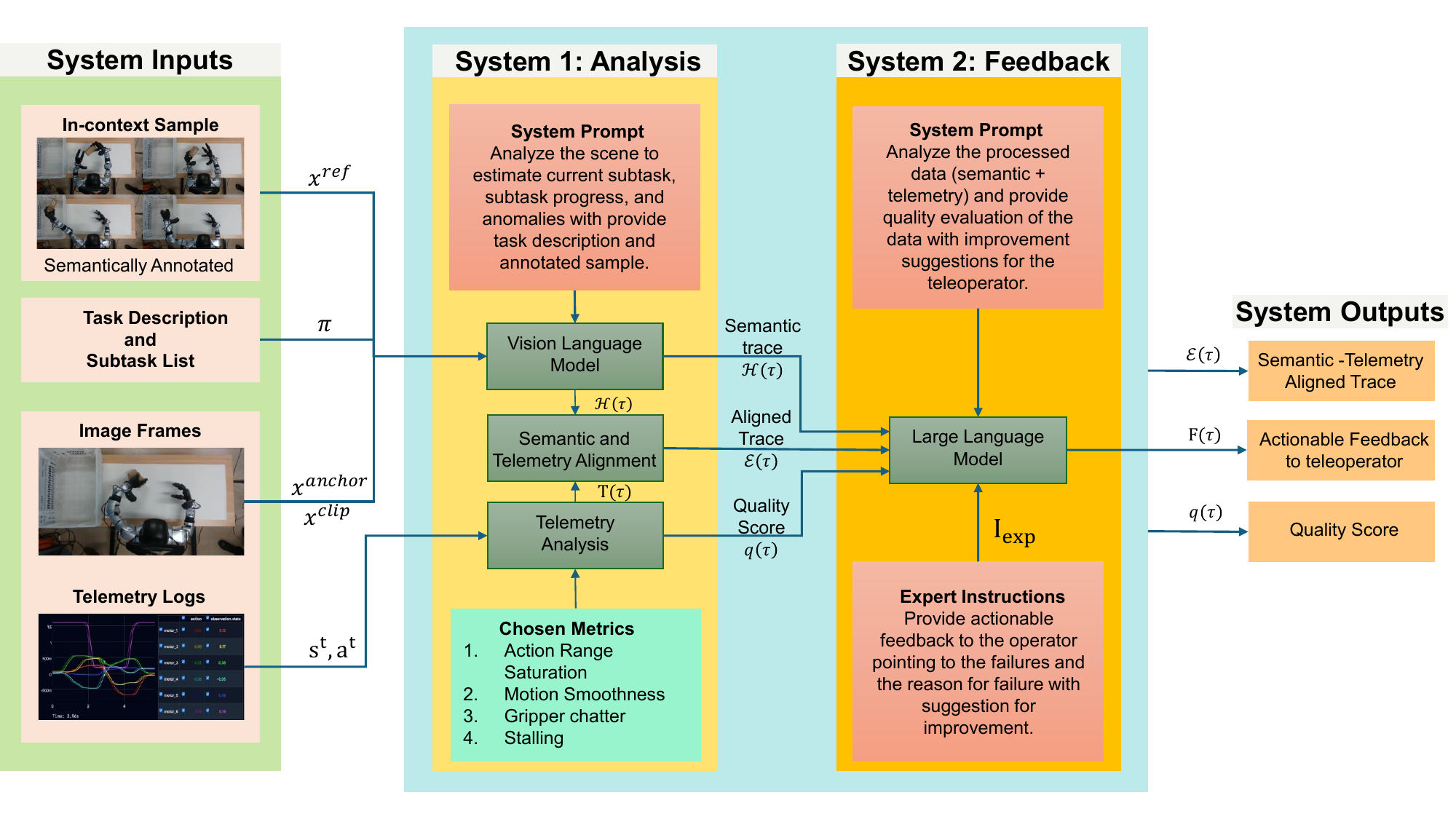}
    \caption{System overview of the proposed DQAF framework for teleoperation. The framework operates in two stages. System 1 analyzes visual observations and telemetry logs, together with task context and reference examples, to infer semantic task progress and compute quality-relevant execution metrics. System 2 then combines these intermediate outputs to generate an episode-level data quality assessment and actionable natural-language feedback for the teleoperator. System prompts and expert instructions provided are for reference.}
    \label{fig:workflow}
\end{figure*}

\section{Methodology}
\label{sec:methodology}

We implement a multimodal pipeline that fuses (i) semantic task progression analysis from a Vision--Language Model (VLM) and (ii) telemetry-based quality analysis, and then (iii) synthesizes operator feedback from temporally aligned data, as shown in Fig.~\ref{fig:workflow}.

\subsection{Pipeline Overview}

Given an episode trajectory $\tau$ and captioned reference frames $x^{\mathrm{ref}}$, the framework produces
$\mathcal{O}(\tau)=\bigl(q(\tau), \mathcal{H}(\tau), \mathcal{F}(\tau)\bigr)$
where $q(\tau)$ is the episode-level quality score, $\mathcal{H}(\tau)$ is the semantic trace, and $\mathcal{F}(\tau)$ is the generated operator feedback. The pipeline consists of four stages:
\begin{enumerate}
    \item Semantic task progress estimation from visual observations
    \item Telemetry quality analysis at segment and episode level
    \item Semantic telemetry alignment with threshold-based diagnosis
    \item Constrained feedback synthesis
\end{enumerate}

\subsection{Semantic Analysis (Plan-Conditioned, Context-Aware)}

Given a task description $d$, we define an ordered subtask plan
$\Pi=\{\pi_1,\ldots,\pi_L\}$.
The plan may be user-provided or automatically generated from the task description.
We perform semantic analysis at a discrete set of update times
$\mathcal{U}=\{t_1,\ldots,t_M\}$.
At each update time $t\in\mathcal{U}$, the vision-language model receives a structured context
\begin{equation}
\label{eq:semantic_inputs}
\mathcal{X}_t=\bigl(x^{\mathrm{anchor}},\,x_t^{\mathrm{clip}},\,\Pi,\,x^{\mathrm{ref}}\bigr)
\end{equation}
where $x^{\mathrm{anchor}}$ is a global anchor frame (e.g., the first image of the episode $I_0$), 
$x_t^{\mathrm{clip}}$ is a short motion clip around time $t$ (e.g., the last five frames $\{I_{t-4},\dots,I_t\}$), 
$\Pi$ is the ordered subtask plan, 
and $x^{\mathrm{ref}}$ are the captioned reference frames from expert demonstrations of the same task.

The model returns a structured output
\begin{equation}
\label{eq:semantic_output}
y_t=\bigl(i_t,\,c_t,\,r_t,\,z_t\bigr)
\end{equation}
where $i_t\in\{1,\ldots,L\}$ is the predicted active subtask index, $c_t\in[0,100]$ is the estimated completion percentage for subtask $\pi_{i_t}$, $r_t$ is the textual rationale, and $z_t\in\{0,1\}$ is an anomaly indicator.

The predicted subtask is constrained to elements of $\Pi$. In our setting, an anomaly ($z_t=1$) includes semantic regressions such as sustained drops in estimated progress over a recent window, repeated backtracking between subtasks, or scene descriptions inconsistent with the expected state for $\pi_{i_t}$.

We convert the local subtask completion estimate into a global progress score
\begin{equation}
\label{eq:global_progress}
p_t=\frac{100}{L}\left((i_t-1)+\frac{c_t}{100}\right)
\end{equation}
so that $p_t\in[0,100]$ represents normalized task progress over the full subtask plan.

This yields the semantic trace
\begin{equation}
\label{eq:semantic_trace}
\mathcal{H}(\tau)=\{h_t\}_{t\in\mathcal{U}},
\qquad
h_t=\bigl(t,\,p_t,\,i_t,\,r_t,\,z_t\bigr)
\end{equation}
where each element records the update time, global progress, active subtask, rationale, and anomaly flag. 

\subsection{Telemetry Quality Analysis} \label{section:telemetry_quality_analysis}

Consistent with our problem formulation, we compute a set of telemetry metrics $\mathcal{M}=\{m_1,\ldots,m_K\}$ over each episode and aggregate them into an episode-level quality score $q(\tau)$ together with segment-level diagnostics. The selected telemetry metrics are motivated by both our initial analysis of collected teleoperation data and prior work on demonstration quality assessment~\cite{sakr2025consistencymatters}. Since these telemetry metrics are reused rather than introduced as novel contributions, we summarize their role here and refer the reader to Forge~\cite{forge} and the original sources for implementation details. In this work, we use four telemetry metrics:

\paragraph{Action Range Saturation.}
Let $\mathbf{a}_t=(a_{t,1},\dots,a_{t,d})$ denote the action at timestep $t$. For each valid action dimension $i$, we define saturation to be when the action lies within a margin $\delta_i$ of its minimum or maximum bound, and compute the overall saturation rate as
\begin{equation}
s_{\mathrm{overall}}=\frac{1}{|\mathcal{V}|}\sum_{i\in\mathcal{V}}\frac{1}{T}\sum_{t=1}^{T} z_{t,i},
\end{equation}
where $z_{t,i}\in\{0,1\}$ indicates whether dimension $i$ is near its bound at time $t$, and $\mathcal{V}$ excludes degenerate dimensions. Larger $s_{\mathrm{overall}}$ indicates frequent limit-hitting behavior and reduced control margin.

\paragraph{Motion Smoothness}
We quantify motion smoothness using Log DimensionLess Jerk (LDLJ) over the joint-state trajectory:
\begin{equation}
\mathrm{LDLJ}(\tau)
=
\log\!\left(
\frac{T_\tau^{3}}{v_{\max,\tau}^{2}}
\int_{0}^{T_\tau}\left\|\dddot{\mathbf{x}}(t)\right\|_{2}^{2}\,dt
\right),
\end{equation}
where $\mathbf{x}(t)$ is the joint trajectory, $\dddot{\mathbf{x}}(t)$ is its jerk, $T_\tau$ is the episode duration, and $v_{\max,\tau}$ is the maximum speed magnitude. Larger LDLJ values indicate jerkier and less stable motion.

\paragraph{Gripper Chatter}
We measure unstable grasp timing by binarizing the gripper command and counting open/close transitions. The chatter rate is defined as
\begin{equation}
\rho_{\mathrm{chat}}=\frac{1}{T_\tau}\sum_{t=1}^{T-1}\lvert b_{t+1}-b_t\rvert,
\end{equation}
where $b_t\in\{0,1\}$ is the binarized gripper state and $T_\tau$ is the episode duration in seconds. High $\rho_{\mathrm{chat}}$ indicates excessive toggling, often reflecting indecision or unstable grasp execution.

\paragraph{Stalling (Static Fraction).}
We estimate stalled execution from the fraction of timesteps with near-zero action magnitude:
\begin{equation}
f_{\mathrm{static}}=\frac{1}{T}\sum_{t=1}^{T}\mathbf{1}\{\|\mathbf{a}_t\|_2<\theta_{\mathrm{static}}\},
\end{equation}
where $\theta_{\mathrm{static}}$ is an adaptive threshold derived from the action distribution within the episode. High $f_{\mathrm{static}}$ indicates prolonged inactivity or stalled control.


\paragraph{Episode-Level Quality Aggregation.}
To match the formulation $q(\tau)=F\bigl(m_1(\tau),\ldots,m_K(\tau)\bigr)$, we map each telemetry metric to a normalized subscore $\phi_k\in[0,1]$ and aggregate them with nonnegative weights $w_k$:
\begin{equation}
\label{eq:episode_quality_aggregation}
q(\tau)=10\cdot\frac{\sum_{k=1}^{K} w_k\,\phi_k\bigl(m_k(\tau)\bigr)}{\sum_{k=1}^{K} w_k}.
\end{equation}
The scalar $q(\tau)$ supports episode ranking/filtering, while the per-metric subscores $\phi_k$, thresholds, and segment flags are passed to the feedback module for localized diagnosis.

\subsection{Segment-wise Metric Analysis} \label{sec:segment_metric_analysis}

We partition each episode $\tau$ of length $T$ into $J$ segments
\begin{equation}
{\tau_1, \dots , \tau_J}, 
\qquad
\tau_j = \{(I_t, s_t, a_t)\}_{t=t^{\mathrm{start}}_j}^{t^{\mathrm{end}}_j}
\end{equation}
where $t^{\mathrm{start}}_j$ and $t^{\mathrm{end}}_j$ denote the start and end timesteps of each segment. 

For each segment $\tau_j$ we compute a set of violation indicators relative to each telemetry quality metric from Section \ref{section:telemetry_quality_analysis}. 
For each metric $m_k \in \mathcal{M}$ we define a threshold $\theta_k$ and compute whether segment $\tau_j$ exceeds (Eq. \ref{eq:exceed_violation}) or is near the threshold (Eq. \ref{eq:near_violation})
\begin{equation} \label{eq:exceed_violation}
    V^E_{j,k} = \mathbf{1}\{m_k(\tau_j) > \theta_k\}
\end{equation}
\begin{equation} \label{eq:near_violation}
    V^N_{j,k} = \mathbf{1}\Biggl\{V^E_{j,k}=0 \land \frac{m_k(\tau_j)-\theta_k}{\max\bigl(\theta_k,\epsilon\bigr)}\Biggr\}
\end{equation}
where $V^E_{j,k} = 1$ indicates that segment $\tau_j$ has exceeded the threshold for metric $m_k$ and $V^N_{j,k} = 1$ indicates that segment $\tau_j$ is near the threshold for metric $m_k$.

\subsection{Cross-modal evidence construction}
We align each segment level violation event $V^E_{j,k}$ and $V^N_{j,k}$ as described in Section \ref{sec:segment_metric_analysis} with the nearest semantic update
\begin{equation}
\label{eq:feedback_nearest_semantic}
 i^{\star}(j)=\arg\min_i \left|t_i-\tfrac{t_j^{\text{start}}+t_j^{\text{end}}}{2}\right|
\end{equation}
This produces a set of aligned evidence items
\begin{equation}
\label{eq:feedback_evidence}
\mathcal{E}(\tau)=\{e_j\}_{j=1}^{N},
\end{equation}
where each $e_j$ summarizes the metric, observed value, threshold, status, time window, and aligned subtask for segment $j$.

\label{subsec:method-feedback}

\subsection{Feedback Synthesis}

We define the feedback-generation input as
\begin{equation}
\label{eq:feedback_inputs}
\mathcal{Y}(\tau)=\bigl(\mathcal{H}(\tau),\,\mathcal{E}(\tau),\,q(\tau),\,\mathcal{I}_{\mathrm{exp}}\bigr),
\end{equation}
where \(\mathcal{H}(\tau)\) is the semantic trace, \(\mathcal{E}(\tau)\) is the aligned cross-modal evidence, \(q(\tau)\) is the episode-level quality score, and \(\mathcal{I}_{\mathrm{exp}}\) encodes expert instructions(text) for how to provide actionable feedback. A constrained LLM then generates
\begin{equation}
\label{eq:feedback_output_contract}
\mathcal{F}(\tau)=\Phi\bigl(\mathcal{Y}(\tau)\bigr)=\{f_{\ell}\}_{\ell=1}^{N_{\mathrm{fb}}},
\end{equation}
where \(f_{\ell}\) denotes the \(\ell\)-th feedback item. The prompt is designed to produce concise, actionable guidance prioritizing task-completion gaps, anomaly and backtracking causes, telemetry threshold violations and near-threshold risks, and corrective actions. Each \(f_{\ell}\) explicitly specifies (i) \textbf{what} failed, (ii) \textbf{where} it failed, and (iii) \textbf{what to change} in the next run.

\section{EXPERIMENTATION AND EVALUATION}

We evaluate the proposed Data Quality Assessment and Feedback framework in a humanoid teleoperation setting through two studies: (1) a diagnostic validation study assessing the reliability of the framework to identify failed or suboptimal episodes, and (2) a pilot user study to examine the effect of immediate post-episode feedback on novice teleoperators.

\subsection{Hardware and Software Setup}

\begin{figure}[t]
    \centering
        \includegraphics[width=0.5\linewidth]{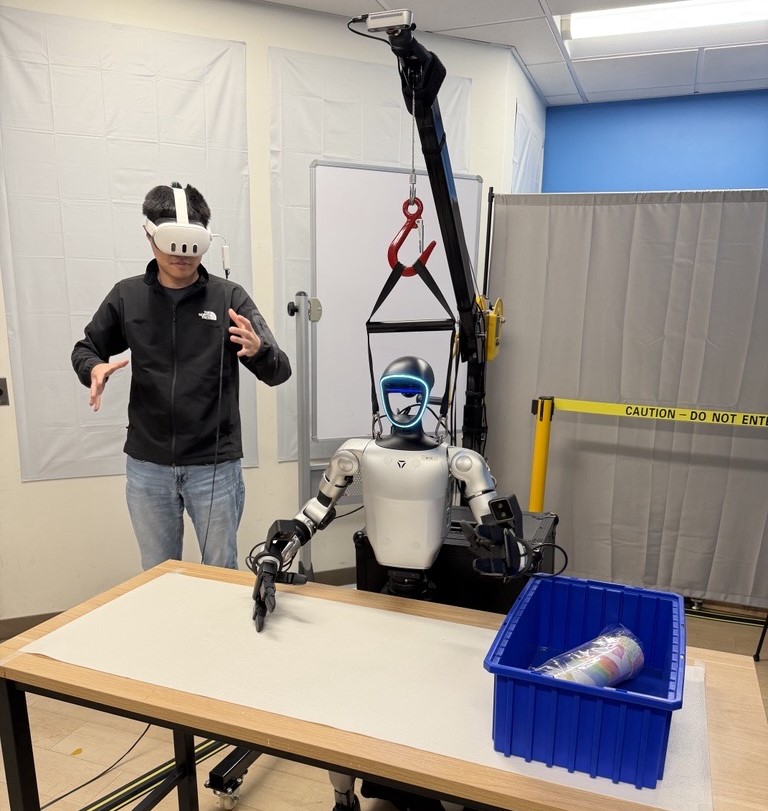}
    \caption{Overview of the experimental setup, showing the Unitree G1 humanoid robot and the teleoperator wearing the XR teleoperation interface}
    \label{fig:setup_overview}
\end{figure}

As shown in Figure~\ref{fig:setup_overview}, the experiments are performed using a Unitree G1 humanoid robot. 
A top-down RGB camera provides workspace observations, while robot telemetry is logged at high frequency, including joint states and commanded actions.
Teleoperation is performed using a Meta Quest-based hand-tracking interface built on the Unitree XR teleoperation stack~\cite{unitreeXR2024}, enabling synchronized recording of multimodal episode data for post-episode quality assessment.

In the implementation used for this study, visual-semantic analysis is performed using Gemini Flash 1.5, while final textual feedback generation is performed using Gemini 3 Pro. 
The framework is model-agnostic and can be instantiated with alternative VLM backends. 
A lightweight Flash-class model was selected for the semantic analysis stage to reduce latency and inference cost when processing multiple image frames, whereas the feedback generation stage operates only on structured textual summaries and therefore permits use of a larger language model. Frames are sampled at 2.5\,s intervals for semantic analysis. 
Metric thresholds are calibrated separately for each task using expert teleoperation demonstrations collected during an initial setup phase, with percentile-based threshold selection applied to the reference data.

\subsection{Experimental Setup and Task Design}

We evaluate the framework on two real-world teleoperation tasks chosen to expose complementary sources of suboptimal demonstration quality, including inefficient motion, unstable grasping, repeated corrective actions, and coordination difficulties in bimanual manipulation.

\textbf{Task~1 (Pick and Place with Box Manipulation).} The robot places a cylindrical object into a box via a bimanual sequence: the left arm positions the box, and the right arm grasps and inserts the object. Success requires the object to be inside the box and both arms to end in a clear, non-interfering configuration.

\textbf{Task~2 (Item Handover and Drop-off).} The robot transfers an object between arms and deposits it into a bin: the right arm picks and presents the object, and the left arm receives, carries, and drops it. Success requires a secure transfer and correct deposit into the bin.

For each task, we curated a small set of expert reference demonstrations that serve as high-quality samples for episode-level assessment.

\subsection{Framework Validation}

Before the pilot user study, we evaluated the DQAF framework as an automated diagnostic tool on a separate dataset of 100 teleoperated episodes which were collected for Task 1. Of these, 72 episodes were successful and 28 were failure cases based on task outcome and post hoc expert review. Each episode was processed by the DQAF pipeline to obtain an episode-level quality assessment and associated diagnostic feedback. The resulting predictions and root-cause descriptions were then compared against expert review using the recorded videos and telemetry logs.

\subsection{Pilot Study Design}

We conducted a pilot user study to examine the utility of immediate post-episode feedback during teleoperation-based data collection. Three novice operators participated in the study. Operator 1 and 3 completed 30 episodes in total for both tasks, while Operator 2 completed 30 episodes of only Task~2. Operator 1 received immediate post-episode feedback generated by the DQAF framework for Task~1 only, Operator 2 recieved no feedback at all, while Operator 3 received feedback after every episode for both tasks. Given the limited number of participants, this study is intended as an exploratory pilot evaluation rather than a definitive controlled comparison.

Each participant first completed a short onboarding session to become familiar with the XR teleoperation interface and basic robot motions. In the immediate-feedback condition, the operator received structured natural-language feedback after each episode, generated from the DQAF analysis described in Section~\ref{sec:methodology}. This feedback summarized the episode quality and suggested corrective actions for subsequent trials. In the no-feedback condition, operators completed all trials without episode-level DQAF feedback during the collection process. All RGB observations, telemetry streams, task outcomes, and DQAF outputs were logged for every episode.

For each episode, we record task success, completion time, the DQAF quality score \(q(\tau)\), counts of detected suboptimalities, and the generated operator feedback for episodes in the immediate-feedback condition.

\begin{figure}[!t]
    \centering
    \includegraphics[width=\linewidth]{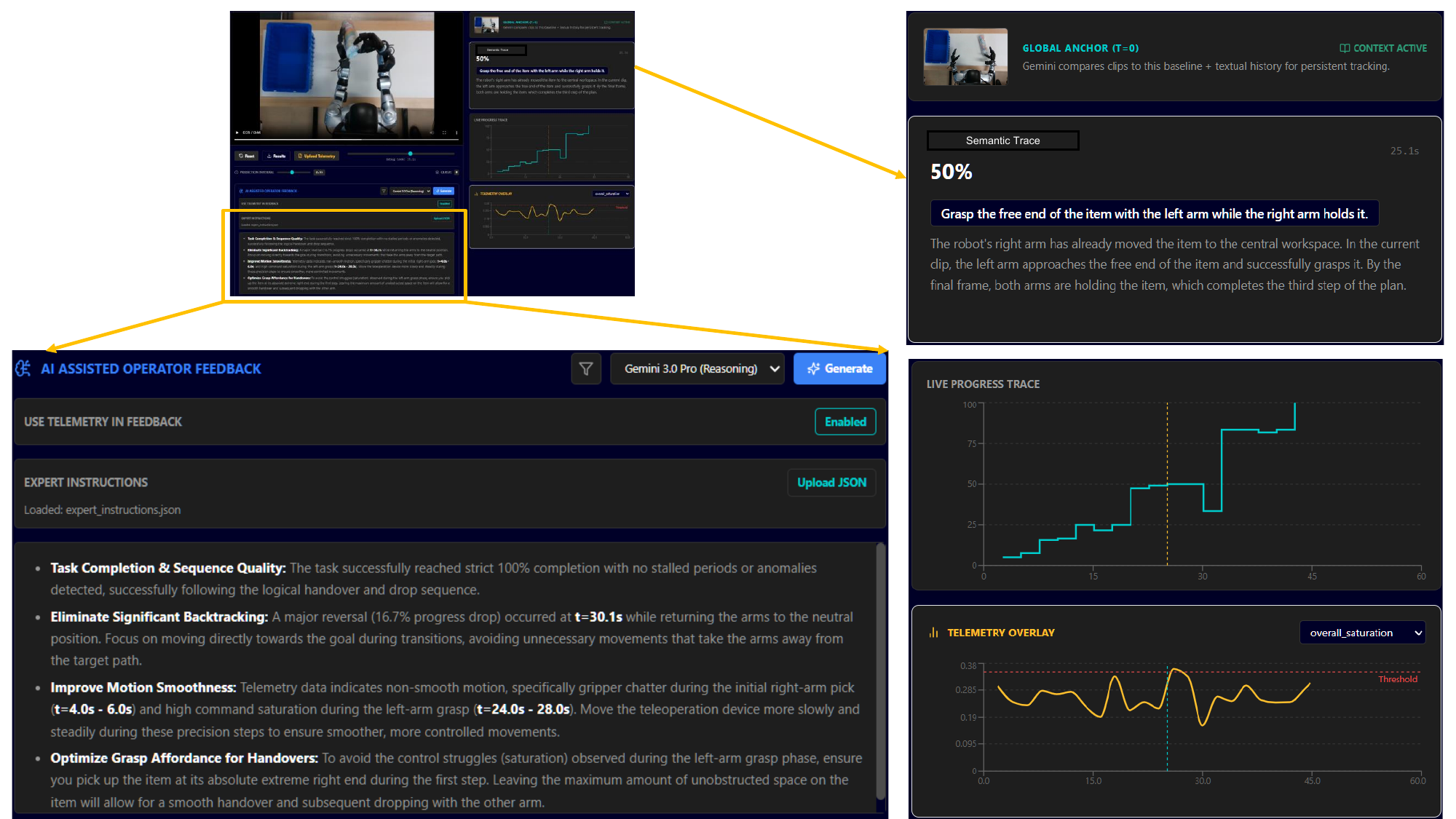}
    \caption{Graphical interface used for DQAF analysis (top left). The Semantic Trace (top right) indicates the inferred subtask and completion progress, the live progress (bottom right) trace visualizes task progression over time, and the telemetry overlay highlights threshold violations in low-level execution metrics. AI assisted Natural Language Feedback (bottom left) for operator based on the semantic trace and telemetry trace}
    \label{fig:framework_validation}
\end{figure}

\section{RESULTS}

\subsection{Framework Validation Results}

The validation dataset comprised 100 episodes (expert review: 72 successes and 28 failures). When assessed with DQAF, the system flagged 26 episodes as failures: 24 were true failures (true positives) and 2 were actually successes (false positives). Conversely, 4 of the 28 failures were misclassified as successes (false negatives), yielding a failure-detection (recall) of 85.7\% (24/28).
This indicates that the framework can identify a substantial fraction of clearly suboptimal demonstrations prior to deployment in the pilot feedback study.

We also measured the latency of the proposed framework. 
Episodes in the validation set were approximately 50\,s long on average and were partitioned into 20 segments of 2.5\,s each.
Semantic trace extraction required 20 API calls per episode, with an average latency of 2.0\,s per call, followed by a final feedback-generation call taking 3.0\,s on average, yielding a total post-episode latency of 43\,s. 
Telemetry analysis took 2.2\,s on average but ran in parallel and did not add to the wall-clock time. 
For comparison, expert human review required 15\,s per episode on average to analyse the video and telemetry logs to provide feedback. When semantic analysis was started during teleoperation, roughly 60\% of the processing completed before episode termination, reducing the remaining post-episode latency to provide feedback to an average 20\,s. This makes the framework practical for immediate operator feedback after the execution.

As illustrated in (Figure~\ref{fig:framework_validation}), the framework identifies suboptimal execution even when the task is completed successfully. 
In the shown episode, the handover is completed, but telemetry trace reveals that both arms become excessively extended during transfer, operating close to joint limits. This issue is not directly evident from the visual outcome alone and is instead detected through the telemetry diagnostics.
The semantic progress trace also shows backtracking during the episode, indicating that after dropping the object in the bin, the left arm hovered randomly before reaching the neutral pose. 
Although the final task outcome is successful, these intermediate behaviors make the demonstration suboptimal, and the episode is therefore flagged by the DQAF framework. 
The generated feedback, shown in the interface, explicitly identifies the relevant time windows and provides concrete suggestions for improvement, in a form similar to expert human guidance.

We also compared the root-cause descriptions produced by the framework against post hoc assessment by an expert teleoperator.
In most cases, the system and expert showed close qualitative agreement on the principal causes of failure, including semantic backtracking, unstable grasping, premature release, stalling, and operation near control or kinematic limits. 
The few cases where the system failed to detect failures occurred when semantic subtask completion could not be reliably inferred from visual observations alone; in particular, partial occlusions and fine-grained progress deviations were judged more accurately by the expert reviewer. 
Conversely, the framework was often more effective at systematically correlating telemetry patterns with semantic events across multiple tracked metrics. 
These results suggest that the framework is most effective as a multimodal diagnostic tool: expert review remains stronger for subtle visual-progress interpretation, while the automated system provides faster and more systematic analysis of telemetry grounded failure modes.

\renewcommand{\dbltopfraction}{0.9}
\renewcommand{\dblfloatpagefraction}{0.8}
\begin{table*}[!tp]
    \small
    \centering
    \caption{\textbf{Summary of teleoperator performance (Mean \(\pm\) Standard Error of the Mean Over 30 Trials)}}
    \label{tab:performance_summary}
    \begin{tabular*}{\linewidth}{@{\extracolsep{\fill}}lcccccc@{}} 
        \toprule
        \textbf{Metric} & \multicolumn{2}{c}{\textbf{Task 1}} & \multicolumn{3}{c}{\textbf{Task 2}} \\
        \cmidrule(lr){2-3} \cmidrule(lr){4-6}
        & \textbf{Op1 (Feedback)} & \textbf{Op3 (Feedback)} & \textbf{Op1 (No Feedback)} & \textbf{Op2 (No Feedback)} & \textbf{Op3 (Feedback)} \\
        \midrule
        Completion Time (s) & $41.6 \pm 2.8$ & $49.2 \pm 1.3$ & $45.4 \pm 2.2$ & $71.8 \pm 6.7$ & $51.0 \pm 2.9$ \\
        DQAF Quality Score & $8.8 \pm 0.5$ & $8.8 \pm 0.4$ & $8.2 \pm 0.3$ & $5.6 \pm 0.6$ & $9.4 \pm 0.2$ \\
        Major Error Count & $1.0 \pm 0.2$ & $1.25 \pm 0.02$ & $2.0 \pm 0.3$ & $2.9 \pm 0.4$ & $1.4 \pm 0.2$ \\
        Success Rate (\%) & 93.33\% & 80\% & 66.66\% & 53.33\% & 86.66\% \\
        \bottomrule
    \end{tabular*}%
\end{table*}
\subsection{Pilot Study Results}

We present pilot study results illustrating benefits of immediate, automated feedback. Table~\ref{tab:performance_summary} summarizes completion time, DQAF quality score, detected suboptimalities, and task success rate averaged over all 30 trials for each task. Task is successful if total error count is three or less. Values are presented as mean ± standard error of the mean (SEM). The SEM indicates how much the sample mean is likely to vary from the true population mean. A smaller SEM suggests a more reliable estimate. Across both tasks, the immediate-feedback condition showed more favorable trends, including higher task completion, higher episode-level quality scores, and fewer detected suboptimalities. In contrast, the no-feedback condition showed slower and less consistent improvement. These observations suggest that immediate post-episode feedback may help novice teleoperators refine their behavior more quickly during data collection.

\begin{figure}
    \centering
    \includegraphics[width=\columnwidth]{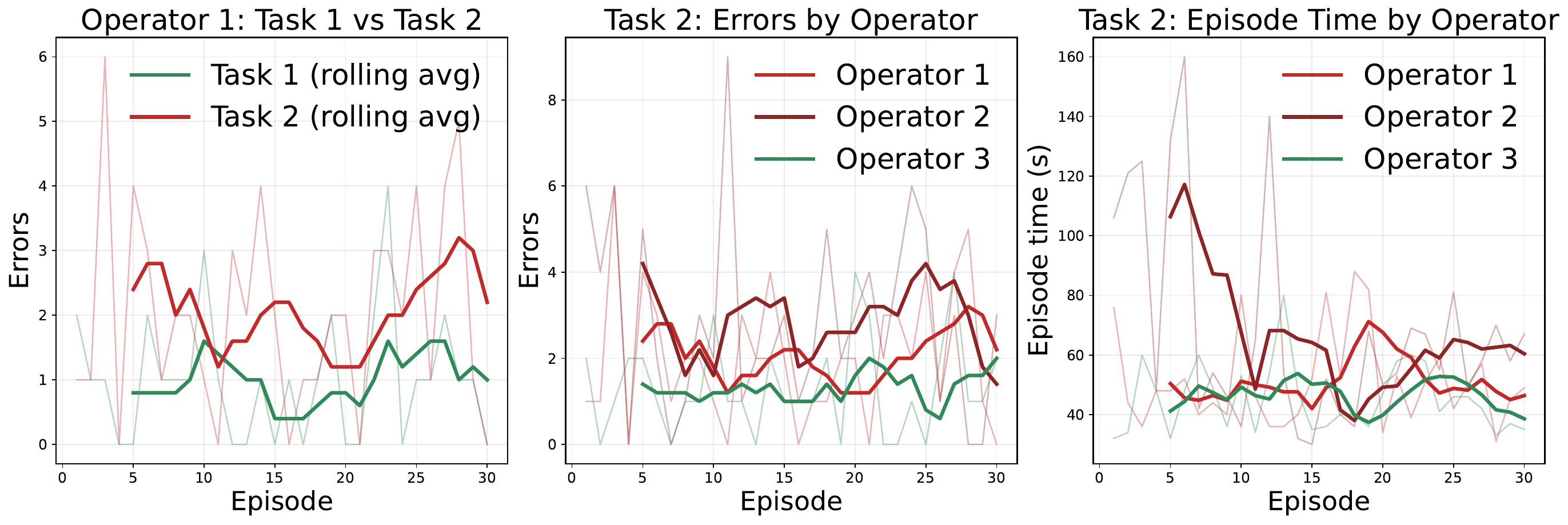} 
    \caption{
        \textbf{(Left)} Operator 1's learning trajectories for Task 1 (Pick-and-Place) with DQAF feedback and Task 2 (Item Handover) without DQAF feedback.
        \textbf{(Middle)} Number of DQAF-identified errors per episode for Task 2 (Item Handover) across three operators.
        \textbf{(Right)} Time taken per episode for Task 2 (Item Handover) across operators.
        For all subplots, bold lines represent 5-episode rolling averages, and faint lines in the background show original episode values. Episodes with no feedback are colored in shades of red, and episodes with feedback are colored in green.
    }
    \label{fig:combined_operator_performance} 
\end{figure}

Figure \ref{fig:combined_operator_performance}  illustrates the learning progression and performance metrics across different operators and feedback conditions. The \textbf{left subplot} shows Operator 1's learning trajectories across two distinct conditions: Task 1 (Pick-and-Place) where DQAF feedback was provided, and Task 2 (Item Handover) where feedback was withheld. A clear difference in learning trajectory is observable as he performs fewer errors on the task where he received immediate feedback from the DQAF system. While direct error comparisons between distinct tasks are inherently limited, the observed differential learning curves for Operator 1, particularly in error reduction, suggest that DQAF feedback improves skill acquisition even when applied to tasks sharing fundamental manipulation challenges. It is also important to note that occasional spikes in error counts can occur due to variations in initial task states across episodes, a common practice in data collection to ensure robust learning. Operator 1 demonstrates improvement in performance (e.g., success rate, DQAF quality score) when receiving DQAF feedback for Task 1. In contrast, performance on Task 2, without the aid of DQAF feedback, shows a slower, less consistent learning curve, often plateauing at a lower level.

The \textbf{middle and right subplots} of Figure \ref{fig:combined_operator_performance} present the number of DQAF-identified errors and the total time taken per episode for Task 2, respectively, across all 3 operators. Operator 3, who received immediate DQAF feedback for Task 2, demonstrates a clear trend of reducing errors and completion time as the number of episodes increased. This suggests that the actionable insights provided by the DQAF framework enabled rapid skill refinement and efficiency gains. In contrast, Operator 1 (without DQAF feedback) and Operator 2 (also without DQAF feedback), show less consistent improvement or higher baseline values for both error counts and completion times. This highlights the challenge for novice operators to self-diagnose and correct complex teleoperation issues without explicit guidance.

\subsection{Qualitative Analysis of Feedback Utility}

The qualitative results show that the framework produces feedback that is both interpretable and actionable. Across low-quality episodes, the system consistently identified recurring causes of failure, including semantic backtracking during handover, unstable initial grasps, premature gripper release, incomplete final reset motions, and abrupt control inputs reflected in telemetry threshold violations. For example, several handover failures were associated with poor grasp affordance and repeated progress reversals, with feedback recommending that the object be grasped at its extreme edge to leave sufficient space for the receiving arm. In other episodes, the system linked failed handovers to premature release and operation near joint limits. Conversely, successful episodes were characterized by monotonic task progression and the absence of anomalies, stalling, or backtracking, while still flagging minor completion issues when the episode was terminated before both arms fully returned to neutral. These examples show that the framework provides more than a scalar score: it translates semantic and telemetry evidence into concrete corrections for subsequent trials.

\section{CONCLUSION }

This work is a first step toward closed-loop, quality-aware teleoperation that provides timely, actionable feedback to teleoperators so they can rapidly improve their teleoperation skills and consistently produce higher-quality demonstrations. 
Improving demonstration quality at the source can directly reduce downstream manual curation and increases the availability of higher quality data.
In turn, better training data can lead to more reliable downstream policy performance and help remove a key bottleneck to broader adoption of Physical AI Systems on the factory floor.

Realizing this potential, however, will require addressing the limitations of this work. 
For example, the semantic analysis stage remains sensitive to partial occlusions, ambiguous subtask boundaries, and imperfect progress estimation, while the telemetry module depends on thresholds calibrated from expert demonstrations and may benefit from more adaptive or learned formulations. 
The current system also provides feedback only after each episode, rather than during execution, which limits its ability to intervene online. In addition, the pilot study is preliminary and should be extended to assess the framework across a broader range of operators, expertise levels, tasks, embodiments, and teleoperation interfaces. Future work will address these limitations through more robust semantic tracking, adaptive telemetry-based quality assessment, real-time in-episode feedback, and direct evaluation of downstream learning performance.


\end{document}